# NeuDiff Agent: A Governed AI Workflow for Single-Crystal Neutron Crystallography

Zhongcan Xiao[a], Leyi Zhang[a,b], Guannan Zhang[c], Xiaoping Wang[a,*]

[a] Neutron Scattering Division, Oak Ridge National Laboratory, Oak Ridge, TN 37831, USA

[b] Department of Linguistics, University of Illinois Urbana-Champaign, Urbana, IL 61801, USA

[c] Computer Science and Mathematics Division, Oak Ridge National Laboratory, Oak Ridge, TN 37831, USA

**Synopsis** NeuDiff Agent accelerates TOPAZ beamtime-to-publication readiness by producing checkCIF-validated structural models through governed tool execution, auditable provenance, and recorded decision rationale.

**Abstract** Large-scale facilities increasingly face analysis and reporting latency as the limiting step in scientific throughput, particularly for structurally and magnetically complex samples that require iterative reduction, integration, refinement, and validation. To improve time-to-result and analysis efficiency, NeuDiff Agent is introduced as a governed, tool-using AI workflow for TOPAZ at the Spallation Neutron Source that takes instrument data products through reduction, integration, refinement, and validation to a validated crystal structure and a publication-ready CIF. NeuDiff Agent executes this established pipeline under explicit governance by restricting actions to allowlisted tools, enforcing fail-closed verification gates at key workflow boundaries, and capturing complete provenance for inspection, auditing, and controlled replay. Performance is assessed using a fixed prompt protocol and repeated end-to-end runs with two large language model backends, with user and machine time partitioned and intervention burden and recovery behaviors quantified under gating. In a reference-case benchmark, NeuDiff Agent reduces wall time from 435 minutes (manual) to 86.5 ± 4.7 to 94.4 ± 3.5 minutes (4.6–5.0× faster) while producing a validated CIF with no checkCIF level A or B alerts. These results establish a practical route to deploy agentic AI in facility crystallography while preserving traceability and publication-facing validation requirements.

## 1. Introduction

In large-scale neutron and X-ray scattering facilities, scientific throughput is increasingly constrained by instrument-to-publication latency(Musslick et al., 2025, Kinhult et al., 2025), namely the time required to transform beamtime into validated, publication-ready results. For single-crystal neutron diffraction on TOPAZ at the Spallation Neutron Source (Coates et al., 2018), the reduction and analysis workflow is well-established and routinely executed(Jørgensen et al., 2014), yet complex structurally and magnetically complex problems(Fancher et al., 2018, Wang et al., 2025) legitimately require many decision points and iterative cycles across configuration (Zikovsky et al., 2011), Mantid-based reduction (Godoy et al., 2020) and integration (Schultz et al., 2014), refinement (Sheldrick, 2015), and publication validation (Spek, 2020). Latency grows when intermediate artifacts, assumptions, and validation outcomes are hard to track, reproduce, and audit across tools and stages, turning necessary iteration into avoidable rework.

To reduce this latency while preserving facility governance expectations, we introduce NeuDiff Agent, a research prototype that accelerates end-to-end TOPAZ single-crystal neutron diffraction analysis from neutron data to validated structures under explicit workflow state, constrained tool execution, and fail-closed verification gates. NeuDiff Agent restricts actions to allowlisted functions, blocks progression when checks fail, and records a complete provenance bundle for every run to support inspection and controlled replay.

Acceleration is only defensible when it is governed. Any AI system that claims to speed crystallographic analysis must ensure that (i) actions are constrained to approved tools, (ii) verification is fail-closed at workflow boundaries, and (iii) decisions and outcomes are auditable and replayable. NeuDiff Agent is designed to preserve scientific judgment by concentrating expert intervention at explicit checkpoints, rather than embedding judgment in opaque automated steps or dispersing it across repeated bookkeeping.

Large language models (LLMs) have made natural-language interfaces to scientific tooling practical (Lu et al., 2024, Gridach et al., 2025, Ren et al., 2025, Arslan et al., 2024), and retrieval-augmented generation (RAG) can reduce ungrounded responses by anchoring outputs in curated context(Lu et al., 2024, Arslan et al., 2024, Minaee et al., 2024). However, facility deployment demands stronger operational guarantees than most general-purpose agent frameworks provide(Gridach et al., 2025, Otoum & Elkhalili, 2026), including enforceable constraints on tool invocation, verification at workflow boundaries, and provenance records that enable inspection and controlled replay rather than post hoc reconstruction.

The key challenge is not generating suggestions, but producing publication-grade crystallographic artifacts through a controlled, replayable execution trace. NeuDiff Agent is designed to meet this requirement directly and makes four contributions: (i) state-aware orchestration that encodes TOPAZ analysis as an explicit, inspectable state model with typed variables spanning data access, reduction, integration, refinement, and validation; (ii) constrained execution with fail-closed verification gates that require schema validity, crystallographic cross-consistency, and tool-verified outcomes before state transitions; (iii) provenance-first audit and replay, producing per-run bundles that capture prompts and configuration, state summaries, dialogue, tool-call arguments and outputs, warnings, exit codes, and timestamps sufficient to reconstruct the analysis path; and (iv) controlled evaluation using a fixed prompt list and repeated end-to-end runs with two LLM backends to quantify time-to-result and document failure modes while maintaining traceability and user accountability.

NeuDiff Agent formalizes the analyst's workflow as an explicit sequence of states with conservative transitions so that routine execution is reliable and deviations are surfaced early. In practical terms, it is designed to reduce avoidable rework by (a) enforcing structured inputs and parameter consistency, (b) restricting tool invocation to approved functions with schema validation, and (c) requiring explicit user authorization at decision points where expert judgment is essential. This design aligns with facility governance needs by making actions inspectable, failures recoverable, and responsibility unambiguous.

A single representative TOPAZ dataset is used as a prototype reference case to demonstrate how a governed, provenance-first agent can accelerate the end-to-end instrument-to-validated-structure workflow. The reported timing and validation outcomes quantify end-to-end acceleration under controlled replay for this dataset and document the audit trail generated by NeuDiff Agent.

A central design choice is that the product of an agent run is not only a final CIF, but also an execution record that supports oversight and reproducibility. For each stage, NeuDiff Agent records tool-call arguments, tool outputs and status, together with workflow-state transitions and gate outcomes. This provenance-first record enables controlled replay for debugging and method verification, and provides a concrete basis for understanding why a result was accepted, corrected, or blocked. In this sense, NeuDiff Agent treats validation as a first-class workflow component rather than an informal final inspection step.

We evaluate NeuDiff Agent in an end-to-end TOPAZ analysis workflow with fixed prompts and repeated runs, comparing a reference manual run against agent-guided execution. The evaluation emphasizes operational metrics that matter in facility settings, including wall-time reduction, intervention burden, and recovery behavior when gates detect inconsistencies. We also assessed whether the final outputs satisfy publication-facing validation expectations, using checkCIF outcomes as a structured endpoint while retaining user accountability for scientific decisions.

The remainder of this paper is organized as follows. We first define the NeuDiff workflow model, including the state representation, allowlisted tool interface, and fail-closed verification gates that govern progression between reduction, integration, refinement, and validation. We then specify the provenance bundle produced for each run and the conditions required for controlled replay and audit. Next, we present a controlled reference-case evaluation that quantifies time-to-validated-CIF, intervention frequency, and failure modes while enforcing a checkCIF acceptance endpoint. We close by discussing practical deployment considerations for facility environments, limitations of the current prototype, and the implications of provenance-first, fail-closed agents for accelerating neutron diffraction analysis without compromising scientific accountability.

## 2. Methods

NeuDiff Agent is a facility-grounded AI agent for TOPAZ single-crystal neutron diffraction. Its primary objective is to accelerate the full analysis path from instrument outputs to validated structures while enforcing facility-grade governance. It combines explicit workflow state, a neutron knowledge schema, constrained tool execution, and fail-closed verification gates, with a provenance bundle produced for each run, as shown in Fig. 1.

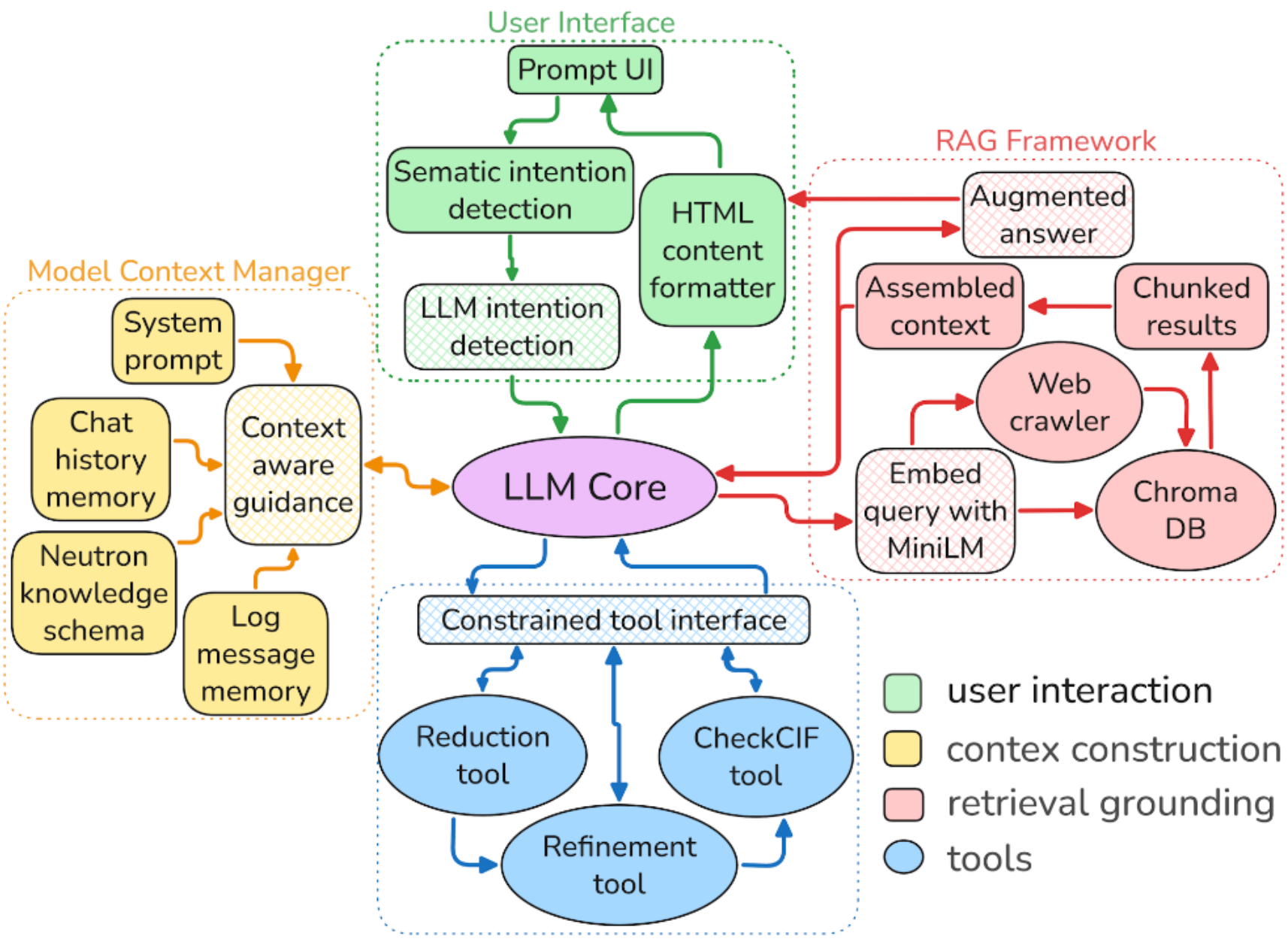

**Figure 1** Logical architecture and information flow of NeuDiff Agent for TOPAZ single-crystal neutron diffraction. User queries are interpreted by the user interface and passed to the LLM core, which is conditioned by a model context manager (system prompt, neutron knowledge schema, chat history, and tool-call logs). Facility documentation is incorporated through a retrieval module that assembles augmented context from curated sources. Scientific actions are executed through a constrained tool interface that invokes reduction, refinement, and checkCIF validation, with inputs and outputs logged for audit and replay.

NeuDiff Agent supports TOPAZ data reduction by assisting users in generating and validating inputs for Mantid-based tool (TOPAZ ReductionGUI) prior to execution, which is a single configuration file (.config) that defines the reduction instance, including run selection, wavelength and resolution limits, peak selection and integration settings, UB initialization or reuse, masking and calibration inputs, and absorption model parameters. NeuDiff Agent checks these inputs for physical plausibility and cross-consistency, proposes targeted corrections when needed, and records the finalized configuration as the provenance record for deterministic reruns and controlled reprocessing. NeuDiff Agent then executes the standard TOPAZ Mantid-based reduction scripts using the saved configuration and exports wavelength-resolved HKLF 2 reflection files for downstream refinement and checkCIF validation. A complete technical description of the workflow, parameter classes, outputs and acceptance checks is provided in Supporting Information S1.

### 2.1. LLM Core

The LLM core interprets user queries, selects appropriate tools and synthesizes explanations. NeuDiff Agent does not treat the model as an unconstrained text generator. Instead, it constrains model behavior within a LangGraph state machine: the model receives a structured prompt comprising (i) a system message defining its role and allowed actions, (ii) the neutron knowledge schema and relevant log messages, (iii) recent chat turns, and (iv) retrieved content from the RAG framework, when applicable. Based on this context, the model either responds in natural language or emits a structured function call.

Tool invocation is explicit and auditable. When the model calls a function, LangGraph packages the current workflow state and the proposed arguments, executes the corresponding Python wrapper around the underlying scientific software, and appends tool inputs, outputs and status codes to the log-message memory. This separation allows different LLM backends to be used without modifying orchestration logic: all backends interact with the same tool interface, the same retrieval document set, and the same memory structures. LLM parameters are summarized in Supporting Information, Table S1 and more backend details are in S2.

### 2.2. User Interface

The user interface is a browser-based front end connecting users to the underlying agent and TOPAZ tools (Fig. 2). It is implemented using the NOVA framework (Jourdain et al., 2025) and built on the Trame library, which provides a server-client structure that separates user-facing components from agent execution. The interface accepts free-text prompts, streams incremental responses from the agent, and exposes structured controls for common actions such as launching a reduction, rerunning integration with modified parameters, or requesting visualizations. It also supports tabular rendering of structured artifacts (for example, reduction configurations and key outputs) and figure generation for analysis results.

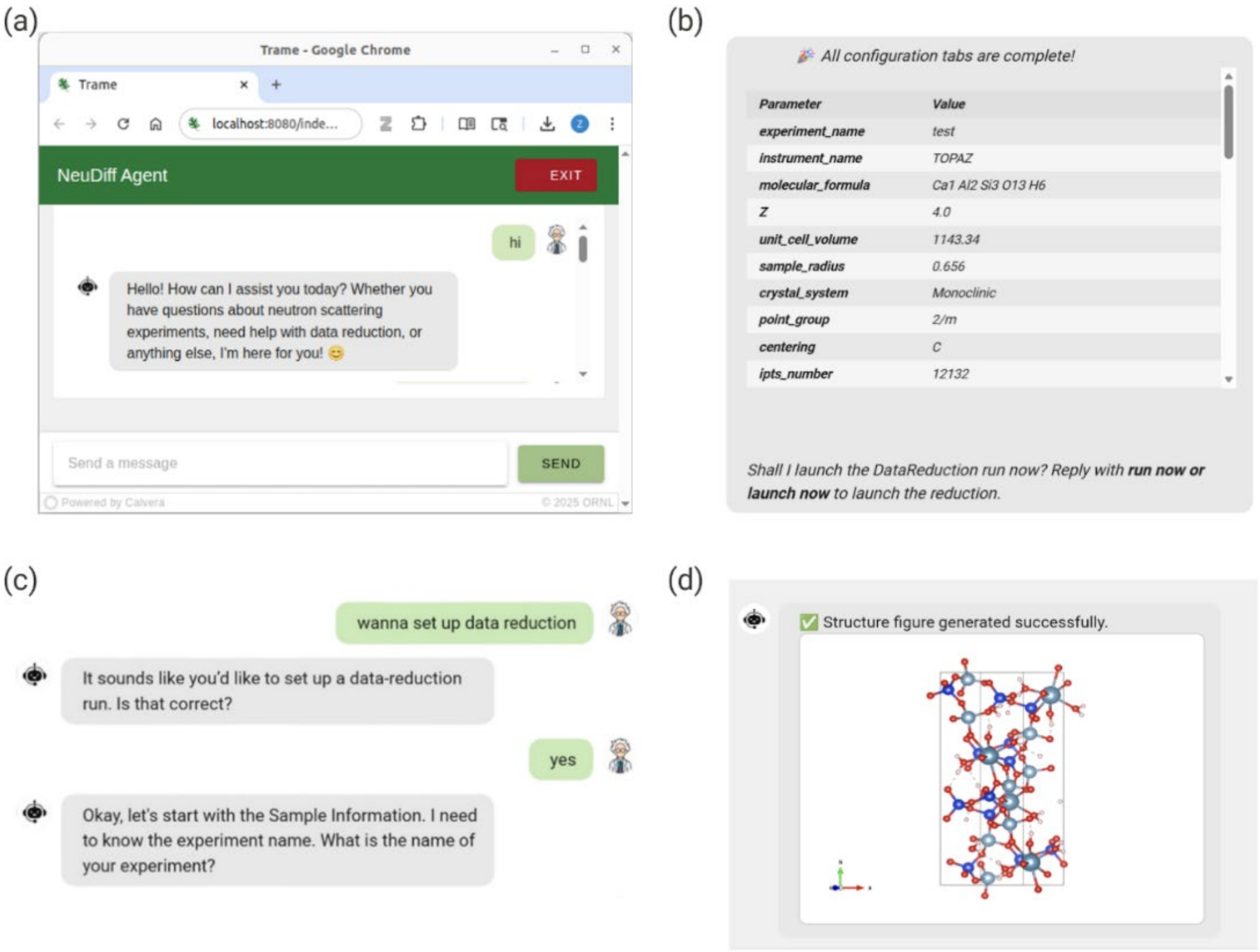

**Figure 2** User Interface (UI) functions of NeuDiff Agent. (a) Browser-based application implemented using NOVA on Trame. (b) Tabular rendering of structured artifacts, including reduction configuration files. (c) Dialogue interface for questions, diagnostics and tool-backed actions. (d) Structure figure generation and rendering from refinement and validation outputs.

User intent is interpreted using a multi-level pipeline. A lightweight semantic layer performs keyword- and pattern-based routing into broad categories (for example, data access, reduction configuration, refinement diagnostics or general questions). For ambiguous or complex requests, the LLM refines the classification and maps the request onto a small set of high-level actions (for example, create a reduction configuration, inspect a UB matrix, or summarize refinement warnings). Before any action that can change analysis results or access facility resources, the user interface requests explicit authorization and blocks execution until approval is provided. This design reduces misconfiguration risk while preserving user control at checkpoints.

### 2.3. Model Context Manager

NeuDiff Agent reduces ungrounded responses by explicitly controlling the context presented to the model. The model context manager maintains a comprehensive LangGraph state that includes the chat history, the current workflow stage, the neutron knowledge schema and the full log of prior tool calls. This state allows the agent to detect task transitions (for example, moving from reduction to refinement diagnostics), reuse validated intermediate artifacts (for example, a UB matrix), and validate new requests against facility constraints.

The LangGraph state comprises four memory types: (i) the system prompt encoding the agent role, safety rules and allowed tools; (ii) the neutron knowledge schema containing long-lived facility facts such as instrument geometry, valid angle ranges, canonical file locations and typical parameter bounds; (iii) chat-history memory for conversational continuity; and (iv) log-message memory as a verifiable record of tool inputs, outputs and status codes. The workflow is implemented as a state machine that transitions through nodes such as interpret requests, select tools, execute tools, and summarize results, always with access to these memory types. After each tool execution, the resulting log entry becomes part of the context for subsequent decisions.

### 2.4. RAG Framework

Users frequently require facility-specific information that is not reliably available from model pretraining, such as instrument modes, sample-environment constraints, or current beamline documentation. NeuDiff Agent therefore employs a RAG framework that queries a local knowledge base of curated, versioned documents, including TOPAZ ReductionGUI user documentation and parameter references, instrument geometry and operating bounds, facility file system conventions and run-metadata guidance, and validation references such as IUCr checkCIF guidance. The knowledge base is updated offline on a controlled schedule and published as versioned, immutable releases. Runtime retrieval is limited to the selected local release, with full provenance logging (release version, source identifier, URL, retrieval timestamp) to support auditability and exact replay. User experimental data are never added to the retrieval knowledge base.

Documents are segmented into semantically coherent chunks and indexed in a local retrieval store that is refreshed periodically. At query time, candidate chunks are ranked using a hybrid similarity and keyword approach, and the highest-ranking context (within a token budget) is supplied to the LLM together with the user query. Implementation details and benchmark test with different models are provided in Supporting Information S3.

### 2.5. Analysis Tools

NeuDiff Agent's agentic behavior derives from its ability to invoke external scientific software in a controlled and auditable manner. The function-call pipeline exposes a small set of high-level functions that wrap TOPAZ domain tools, including Mantid-based scripts for data access and reduction, refinement tools such as SHELXL (Sheldrick, 2015), IUCr checkCIF (Spek, 2020) validation, and structure-visualization utilities such as VESTA (Momma & Izumi, 2011). Each function is designed to be side-effect-aware and idempotent where possible: configurations are validated prior to execution, inputs and outputs are logged, and intermediate files are written into an experiment-specific workspace.

To translate model outputs into defensible workflow actions, NeuDiff Agent applies verification gates that fail-closed. Inputs and intermediate results are checked against instrument constraints, cross-input consistency, and tool-verified outcomes. When a check fails, the agent blocks progression and requests correction rather than proceeding with an unverified step (Table 1).

**Table 1** Verification gates and constraints used by NeuDiff Agent.

| Constraint class | What is checked (examples) | Agent action when check fails |
| --- | --- | --- |
| Hard bounds and schema checks | Unit cell volume and metric sanity; wavelength/d-spacing and resolution range; required file/run formats. | Rejects the value, explains the violation, and proposes an acceptable range or format before rerunning. |
| Cross-input consistency checks | Space group vs cell and symmetry; UB matrix and | Stops the workflow, highlights the inconsistent inputs, and requests |
| | orientation metadata; run list and calibration/background compatibility. | corrected metadata or regenerated orientation. |
| Tool-verified execution checks | Mantid reduction log signatures and output completeness; SHELXL convergence and residual patterns; displacement parameter plausibility. | Flags the failure mode and suggests a targeted corrective step (parameter change, restraint, or data selection) with justification. |
| Publication validation checks | CIF metadata completeness; checkCIF alerts (A/B must be cleared); recorded change log of edits and rationale. | Enters a validation loop, applies metadata or justified refinement updates, and records what changed and why. |

A typical analysis proceeds as follows. The user selects a dataset on the SNS analysis servers. The agent proposes a reduction configuration and invokes the reduction wrapper. After reduction completes, the agent inspects logs and warnings and may propose follow-up actions such as re-indexing with an updated UB matrix or adjusting masking and integration parameters. Once reduction outputs meet minimal acceptance checks, the agent proceeds to refinement via a high-level function call that generates and executes a SHELXL input, while requesting user input when scientific judgment is required. Finally, the resulting CIF is validated using the IUCr checkCIF tool and corrected in a controlled loop until publication-facing validation criteria are met. After each stage, NeuDiff Agent summarizes outcomes, highlights any gate failures, and requests explicit user authorization before executing actions that alter results.

### 2.6. Constraints and verification gates

NeuDiff Agent treats quality control as an explicit workflow stage rather than an implicit, ad hoc activity. Before initiating expensive computation, it validates proposed configurations and evaluates tool outputs against conservative rules intended to prevent common silent failure modes that otherwise propagate and waste analyst time. Concretely, NeuDiff Agent applies (i) hard bounds on unit-cell parameters, wavelength, d-spacing, and resolution limits; (ii) consistency checks on UB matrices and orientation metadata; (iii) experiment-context checks on file paths and run identifiers to prevent cross-experiment contamination; (iv) parsing of Mantid and SHELXL logs for known failure signatures, including non-convergence and physically unreasonable displacement parameters; and (v) workflow gating that blocks refinement and CIF validation until prerequisite artifacts are present and minimal checks have been satisfied (Table 1).

To make the governance model explicit, we define a verification gate as a deterministic check that must pass before a workflow state transition is accepted. Gates fail-closed: if a check fails, the workflow halts and requires a user-authorized corrective action. We refer to each such action as an intervention event. Gate outcomes and intervention events are recorded together with tool-call arguments, tool outputs, and timestamps in a per-run audit directory (see Supporting Information S4).

### 2.7. Security and Data Governance

Tool execution (for example, Mantid scripts, SHELXL, VESTA and checkCIF submission) is performed only through explicit function calls with user authorization at execution checkpoints. Remote execution is restricted to SNS analysis file systems and user-owned directories. Scientific data remain on SNS storage, and only derived artifacts (for example, plots, HKL files and CIFs) are exposed through the web interface for download. Credentials (for example, SSH keys and model API tokens) are stored as host environment variables and are not injected into prompts or run logs. Tool-call arguments, outputs and status codes are logged to support audit and controlled replay. More details are discussed in Supporting Information S5.

## 3. Results and Discussion

We evaluate NeuDiff Agent as a governed workflow executor prototype with two facility-relevant objectives: (1) accelerate end-to-end time-to-validated-structure, and (2) make each step inspectable, replayable and safe under facility governance.

To isolate workflow behavior from prompt drift, we use a controlled protocol with a fixed prompt list, repeated runs and two representative LLM backends. We report time-to-result with partitioned human and machine time extracted from execution logs, and we characterize robustness through failure modes, intervention burdens, and recovery behavior under fail-closed verification gates. The details regarding setup of experiments including software version and hardware models are discussed in Supporting Information S6. In addition to the benchmark test, we also conduct tests for additional material and obtained publication quality CIF results.

### 3.1. Benchmark case study and evaluation protocol

To anchor the benchmark, we selected a TOPAZ scolecite dataset representative of routine operation and used the original X-ray CIF as the starting model. The reference manual run follows the standard reduction, integration, refinement and validation pathway and provides a reference point for time-to-result and the sources of rework that the agent is designed to eliminate. The reference manual run is a single representative experienced-user run on the same workstation and software environment used for the agent-assisted runs; it is used as a reference point for end-to-end speedup, whereas variability and error bars are computed from five repeated NeuDiff Agent runs per backend.

To control task definition across runs, we recorded the complete unaided workflow, including user inputs, intermediate artifacts and decision points, and converted it into a fixed prompt list. Each benchmark run uses the same fixed prompt list, dataset, retrieval release, and pinned tool configuration to control the task definition. The agent may still take different intermediate actions, but all actions remain governed by the same allowlisted tools and fail-closed verification gates. Details are provided in Supporting Information S7: the fixed prompt list used for controlled replay is in S7.1; the workflow stage and types of time cost is discussed in S7.2; the details regarding the crystallographic refinement is in S7.3; the details of CIF validation process are in S7.4. The overall workflow of agent-guided data processing from raw data to validated CIF file, including the user action, agent response, and agent-generated results, are shown in Fig. 3.

Because NeuDiff Agent is intended to support, not replace, expert judgment, agent-assisted runs are explicitly human-in-the-loop at gate boundaries. The user authorizes each tool execution and approves any corrective action required to clear a failed verification gate. Each corrective action is recorded as an intervention event in the execution log, enabling explicit accounting of intervention burden rather than assuming silent recoverability.

We record end-to-end wall-clock time from the first user query to the final CIF and decompose it into machine time, defined as LLM latency plus tool execution, and user time, defined as the remaining wall-clock time. Success is defined by completion to a final CIF that meets publication-facing validation, together with a complete, replayable execution log. Task segmentation and definitions of wall, machine and user time are given in Supporting Information S6.2. In this benchmark, the practical scientific bottleneck for non-expert users is completing the hydrogen substructure that is absent or unreliable in the starting X-ray model;

NeuDiff Agent accelerates convergence by guiding hydrogen placement under explicit verification gates.

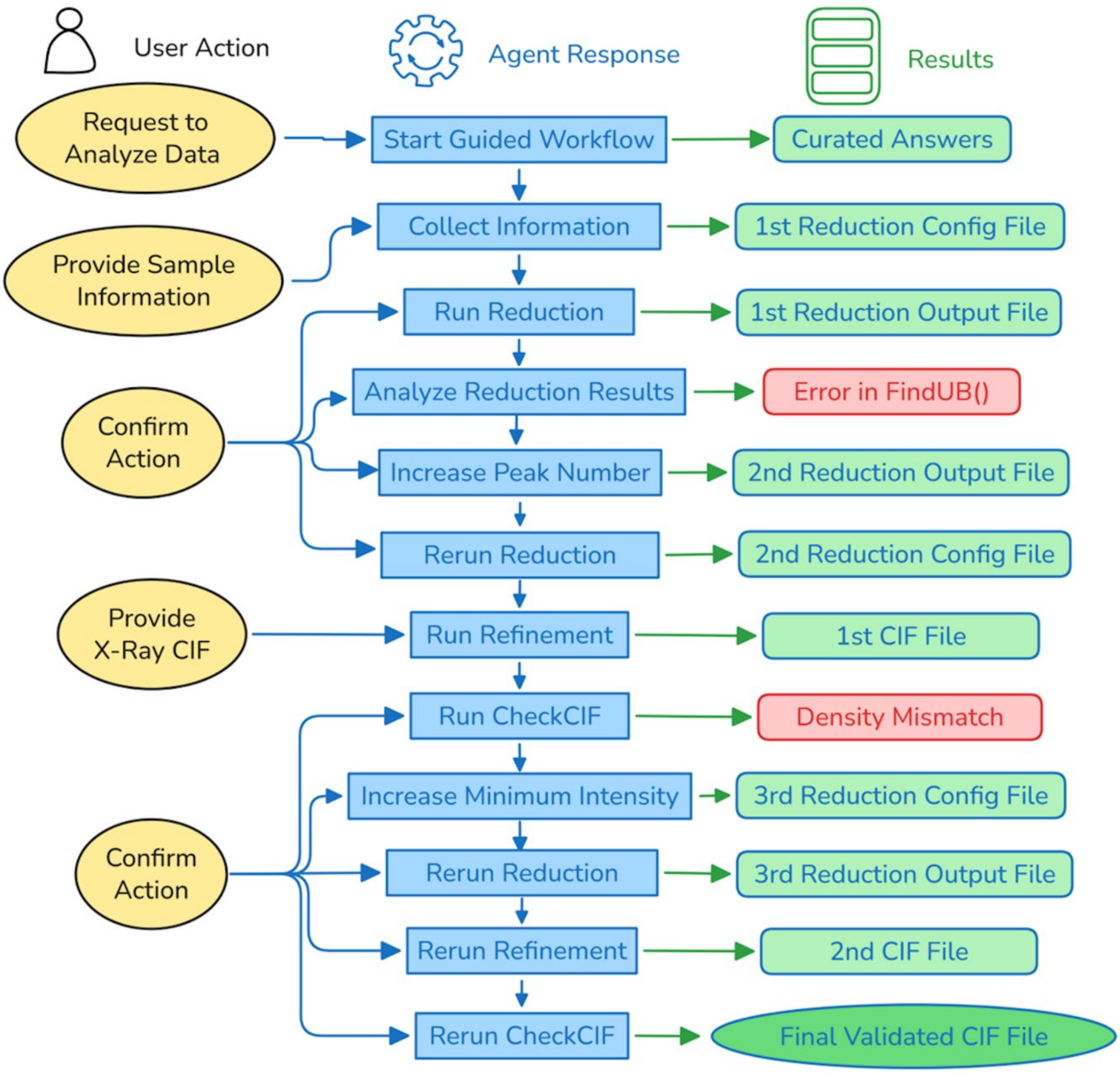

**Figure 3** End-to-end benchmark workflow executed on TOPAZ, represented as auditable agent execution from raw event data and a starting CIF to a validated, publication-facing CIF. User inputs (yellow) initiate execution. Agent states and tool-grounded actions (blue) advance through reduction, refinement and validation under explicit authorization points. Verification gates enforce fail-closed checks (for example, FindUB success, density mismatch, and checkCIF alerts) and trigger corrective loops when criteria are not met. Intermediate configurations, tool outputs and validation outcomes are logged as artifacts to support audit and controlled replay.

To assess robustness across model choices, we evaluated NeuDiff Agent with two representative LLM backends: an online model (Gemini 3.0 Pro) and a locally hosted open model (GPT-OSS 120B). Each backend was run five times as independent end-to-end benchmark runs under a controlled task definition protocol (fixed prompt list, frozen retrieval release, pinned toolchain). This benchmark foregrounds facility reality: correctness depends on many small, instrument-specific decisions and stage-to-stage consistency. NeuDiff Agent does not replace crystallographic judgment. It reduces routine friction by enforcing state-aware routing, validating parameters and metadata before expensive steps, and surfacing stage-specific diagnostics that make recovery fast and auditable.

### 3.2. Failure modes, intervention burden, and recovery behavior

NeuDiff Agent mitigates common workflow failure modes through governed interaction patterns that align with explicit workflow stages (Fig. 4): retrieval-grounded question answering to resolve missing context, parameter validation to prevent invalid configurations, results assessment to detect inconsistent outputs, guided iteration that triggers tool execution only when prerequisites are met, and publication validation that treats validation as an explicit stage. These behaviors reduce intervention burden by converting many latent errors into early, localized gate checks rather than late-stage rework.

A recurring facility pain point is that users must consult instrument and method context while configuring reductions or interpreting refinement output. NeuDiff Agent answers such questions with retrieval-grounded responses and reuses the same context to validate subsequent parameter choices, reducing the risk that guidance remains disconnected from execution. When a gate blocks progression, the agent reports (i) why the check failed, (ii) which inputs or outputs triggered the failure and (iii) a concrete, stage-appropriate next action. This makes recovery attributable to user-approved decisions rather than to implicit agent behavior.

Importantly, NeuDiff Agent does not bypass validation by relaxing refinement quality or suppressing checks; it blocks progression until gates are cleared and records what changed and why. Refinement statistics (Table 2) are reported to demonstrate that, once missing hydrogen positions were located and incorporated into the neutron model, refinement converged to publication-acceptable residuals. More information including reflection counts and theta(max) for the first and final validated CIF are reported in Supporting Information (Table S2). The crystallographic statistics of the final validated CIF are reported in Supporting Information Table S3. The Hydrogen-bond geometry is included in Supporting Information Table S4.

**Table 2** Refinement statistics before and after NeuDiff Agent-guided improvement.

| Metric parameter | First Model[a] (1st CIF file) | Final Model[b] (Final CIF file) |
|---|---|---|
| R1 | 0.1846 | 0.0554 |
| wR2 | 0.4594 | 0.1297 |
| GoF | 2.195 | 1.062 |

[a] First model is derived directly from the X-ray CIF and initial calculations of reduction and refinement. [b] Final model of the validated neutron CIF is from agent guided improvement to the configurations after analyzing the output. R1 is the conventional residual on |F|, wR2 is the weighted residual on $F^2$, and GoF is the goodness-of-fit based on the applied weighting scheme.

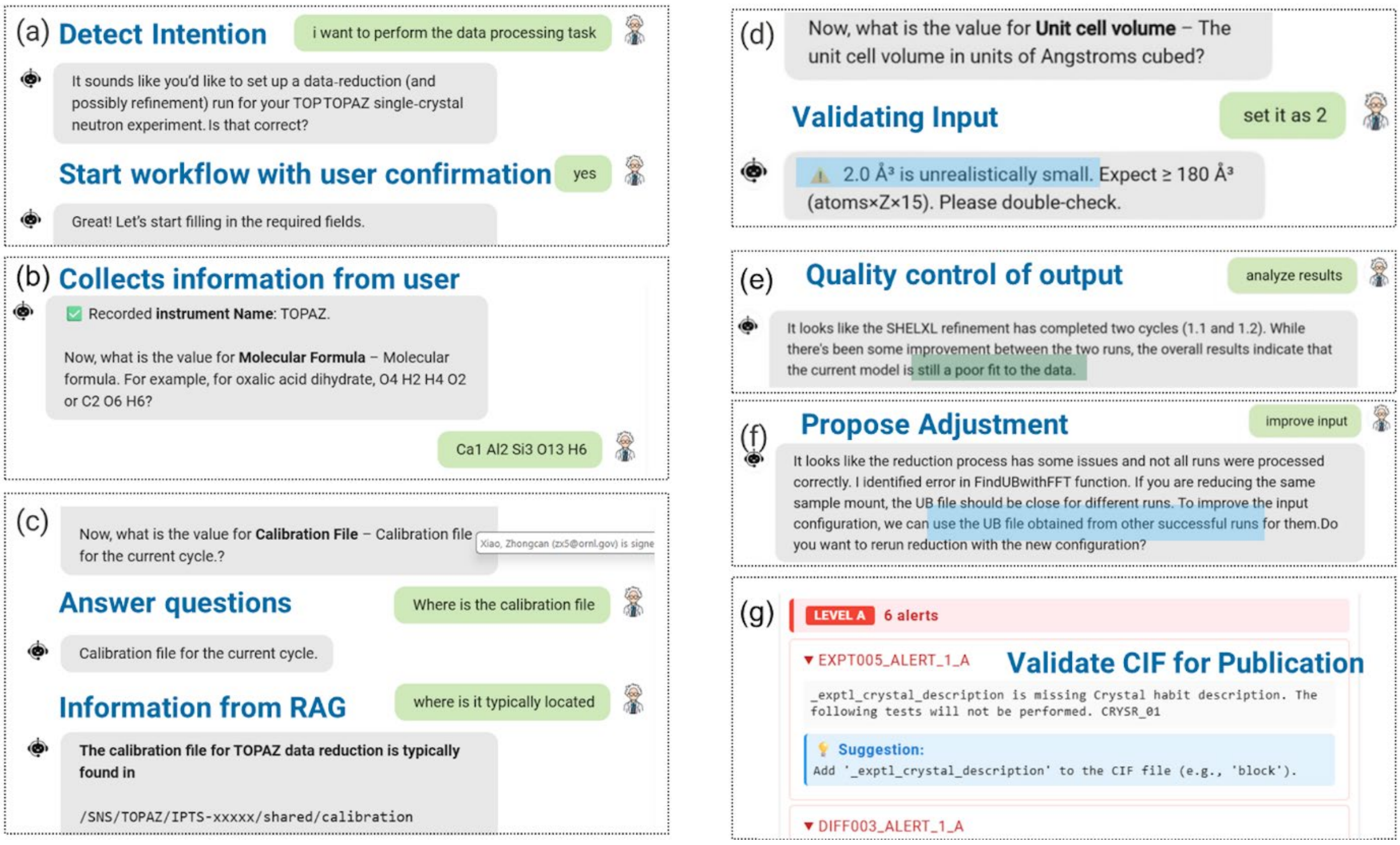

**Figure 4** Representative agent interactions mapped to workflow stages in the benchmark. A typical run proceeds as follows: (a) detect the user's intent to process data and enter the workflow loop; (b) collect required inputs from the user; (c) answer user questions, retrieving supporting information when needed; (d) validate inputs as execution progresses; (e) assess the quality of intermediate outputs; (f) propose and apply adjustments to calculations to address issues detected in prior steps; and (g) validate the final CIF for publication. The emphasis is on stage-specific diagnostics and fail-closed gating that constrain the next permitted action.

### 3.3. Productivity and user effort

We quantify productivity using end-to-end wall-clock time and user effort computed from execution logs. Reported times include user interaction as well as waiting time for both model responses and tool execution, so the benchmark reflects end-to-end latency rather than isolated compute speed.

The time-cost comparison is shown in Fig. 5. End-to-end wall-clock time is measured from the first user query to the final CIF and partitioned into user time and machine time. Across five independent runs per backend, NeuDiff Agent reduces the baseline time of 435 min to 86.5 ± 4.7 min (Gemini 3.0 Pro) and 94.4 ± 3.5 min (GPT-OSS 120B).

The productivity gain is explained by workflow mechanics rather than model verbosity. Verification gates catch configuration and metadata issues before downstream steps and reduce expensive backtracking. Explicit workflow state keeps the next required action unambiguous and concentrates user decisions at checkpoints where judgment is required, rather than dispersing attention across repeated bookkeeping and re-entry of parameters.

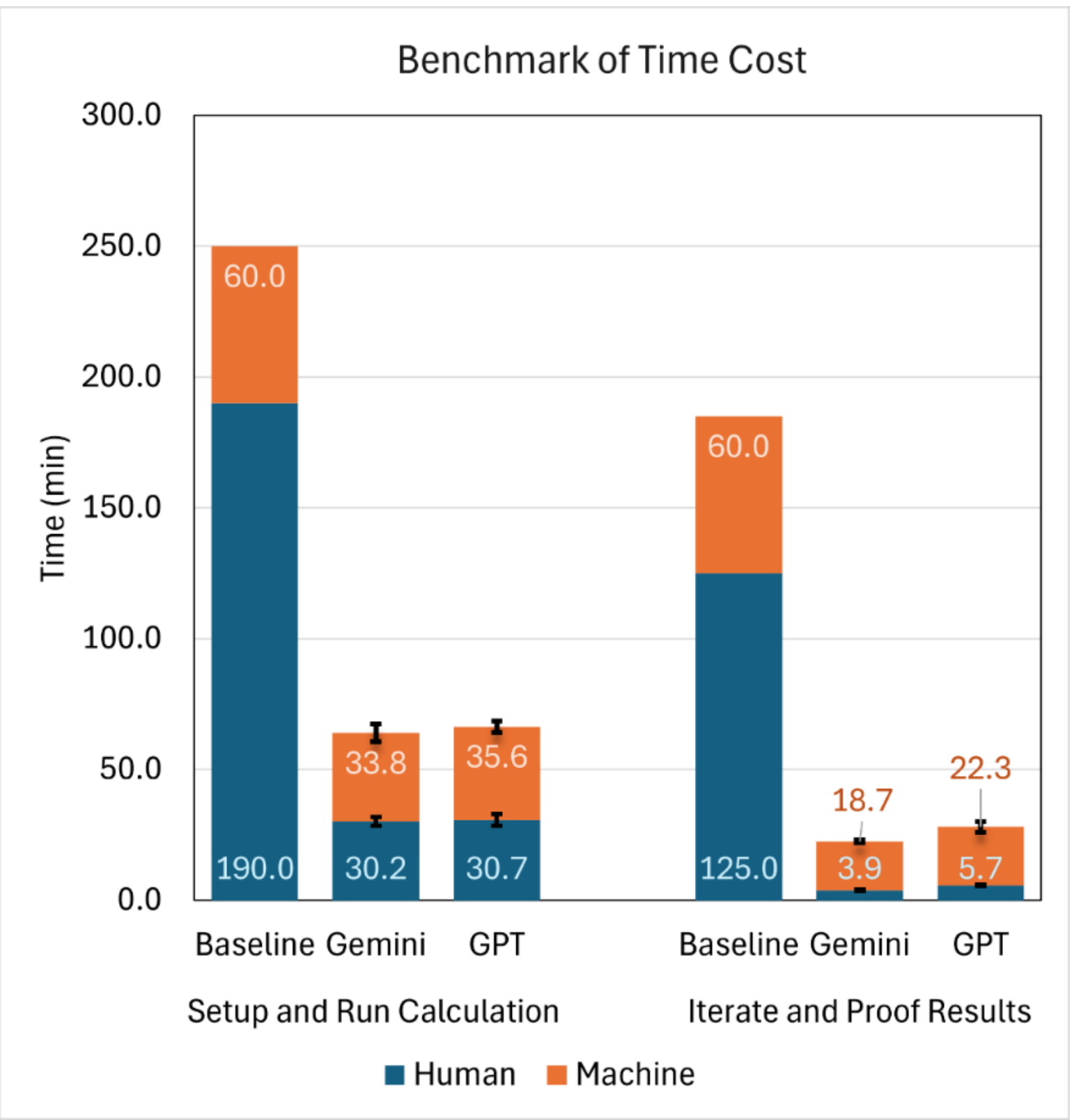

**Figure 5** Wall-clock time benchmark comparing the reference manual run workflow and agent-guided execution. Bars decompose time into user time and machine time, and points indicate run-to-run variability. Values show means and standard deviations across five independent controlled replays per backend.

### 3.4. Publication readiness and CIF validation

Across five independent runs per backend, time-to-result variability was small (standard deviation 3.5-4.7 min across backends), consistent with stable execution under controlled replay. Importantly, stability is not inferred from timing alone. Each run produces an audit trail that supports diagnosis of deviations because tool-call arguments, tool outputs and gate outcomes are recorded.

NeuDiff Agent treats submission readiness as an explicit workflow stage in which refinement outputs are verified against publication-facing criteria. In the scolecite case study, the checkCIF verification gate triggered at the start of publication validation with 6 level A, 0 level B, 5 level C, and 15 level G alerts (Table S5). We resolved all level A alerts, 1 level C alert and 1 level G alert. This required 7 intervention events: 6 metadata completions and 1 chemical-formula consistency correction. The metadata are obtained from reduction step results, and the chemical-formula was changed to reflect IUCr alphabetical order convention. These corrections are all proposed by NeuDiff Agent based on the general crystallography knowledge schema. They are documented in the change log in Table S6. In the final CIF file, all A and B alerts are cleared and leaving only level C and level G items that did not compromise crystallographic interpretation. This accounting makes recovery measurable: interventions are enumerated, attributable to a gate outcome, and paired with the exact CIF fields modified. An illustrated workflow of how the validated CIF file is obtained is shown in Fig. 6.

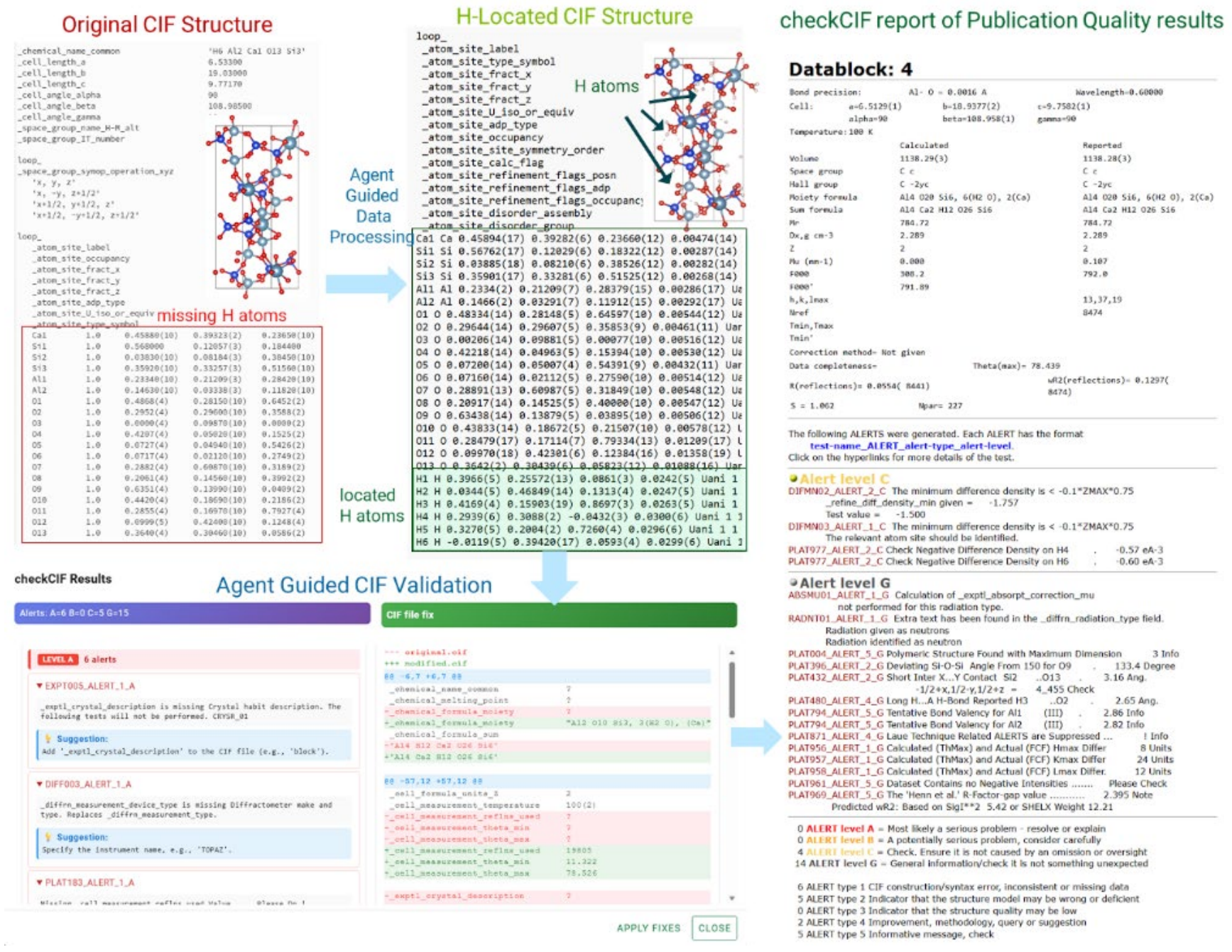

**Figure 6** Outcome validation for the agent-guided workflow. The starting CIF derived from X-ray data is supplied to generate the final neutron-refined CIF through agent-guided pipeline. After hydrogen completion, refinement converged to acceptable residuals and the final CIF was validated for submission, showing a checkCIF report with no level A or B alerts, illustrating a publication-facing acceptance check for the auditable workflow.

## 4. Conclusion

NeuDiff Agent demonstrates a facility-grounded approach for accelerating single-crystal neutron diffraction analysis from neutron data to validated structures. Its core claim is that acceleration is defensible only when paired with explicit governance and complete provenance: NeuDiff Agent constrains actions to allowlisted tools, blocks unsafe transitions with fail-closed verification gates, and produces an audit trail of tool-call arguments, outputs and execution status that makes the full path inspectable and replayable.

In the TOPAZ scolecite benchmark, NeuDiff Agent reduced end-to-end wall time from 435.0 min to approximately 86.5-94.4 min across two LLM backends while producing validated CIF with no checkCIF level A or B alerts. The practical source of this compression is operational: the agent prevents avoidable rework early, shortens iteration cycles between tools, and concentrates user involvement at explicit decision points where expert judgment is required. This shifts effort from bookkeeping toward interpretation while preserving user accountability and facility governance.

For national user facilities, these properties matter because they shorten instrument-to-validated-structure latency without weakening validation standards. NeuDiff Agent reduces avoidable rework by enforcing consistent execution, capturing provenance automatically, and concentrating expert attention on the scientific judgments that require human oversight, with each intervention recorded for governance review.

Broader validation is underway as a facility engineering activity. This manuscript focuses on a single reference-case benchmark that makes the governed execution pattern concrete and reviewable. Extending controlled replay to additional TOPAZ datasets spanning varied crystal systems, experimental conditions, and data quality will quantify how general the current verification gates are and identify failure modes that require new checks or improved tool wrappers.

**Acknowledgements** This research used resources at the Spallation Neutron Source, a DOE Office of Science User Facility operated by the Oak Ridge National Laboratory. We acknowledge research sponsored by the Laboratory Directed Research and Development Program of Oak Ridge National Laboratory, managed by UT Battelle, LLC, for the U.S. Department of Energy, and DOE Office of Science, Advanced Scientific Computing Research, via the AI4Science Program under the grant ERKJ388.

**Conflicts of interest** The authors declare no conflicts of interest.

**Data availability** The NeuDiff Agent source code is available at https://code.ornl.gov/5xw/NeuDiff-Agent. Access to raw instrument data and any facility-internal documentation is subject to SNS data policies.